\pdfoutput=1

\documentclass[11pt]{article}

\usepackage[]{acl}

\usepackage{times}
\usepackage{latexsym}

\usepackage[T1]{fontenc}

\usepackage[utf8]{inputenc}

\usepackage{microtype}

\usepackage{inconsolata}
\usepackage{graphicx}
\usepackage{placeins}
\usepackage{float}

\usepackage{multirow}

\usepackage{algorithm}
\usepackage{algpseudocode}
\usepackage{amsmath}

\newcommand{\cfiltlogo}{\raisebox{3.4pt}{\includegraphics[scale=0.026]{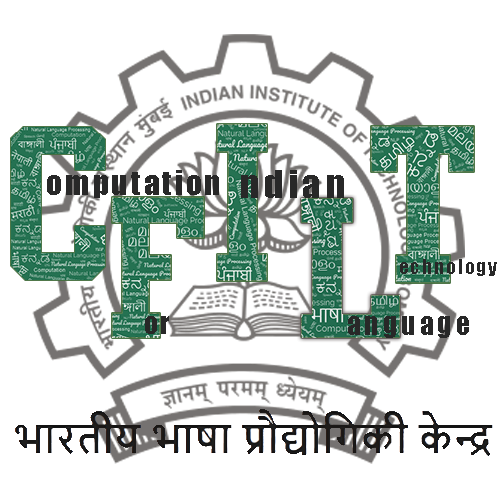}}}
\newcommand{\PAIlogo}{\raisebox{3.4pt}{\includegraphics[scale=0.09]{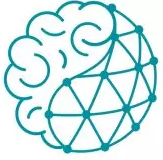}}}

\title{Giving the Old a Fresh Spin: Quality Estimation-Assisted Constrained Decoding for Automatic Post-Editing}

\newcommand*{\affaddr}[1]{#1} 

\newcommand*{\email}[1]{\texttt{#1}}

\author{%
Sourabh Deoghare\cfiltlogo, Diptesh Kanojia\PAIlogo and Pushpak Bhattacharyya\cfiltlogo\\
\affaddr{\cfiltlogo CFILT, Indian Institute of Technology Bombay, Mumbai, India}\\
\affaddr{\PAIlogo Institute for People-Centred AI, University of Surrey, United Kingdom}\\[.2em]
\email{\{sourabhdeoghare, pb\}@cse.iitb.ac.in}, \email{d.kanojia@surrey.ac.uk} \\
}

\begin{document}
\maketitle

\begin{abstract}
Automatic Post-Editing (APE) systems often struggle with over-correction, where unnecessary modifications are made to a translation, diverging from the principle of minimal editing. In this paper, we propose a novel technique to mitigate over-correction by incorporating word-level Quality Estimation (QE) information during the decoding process. This method is architecture-agnostic, making it adaptable to any APE system, regardless of the underlying model or training approach. Our experiments on English-German, English-Hindi, and English-Marathi language pairs show the proposed approach yields significant improvements over their corresponding baseline APE systems, with TER gains of $0.65$, $1.86$, and $1.44$ points, respectively. These results underscore the complementary relationship between QE and APE tasks and highlight the effectiveness of integrating QE information to reduce over-correction in APE systems.
\end{abstract}

\section{Introduction}
Automatic Post-Editing (APE) focuses on developing computational approaches to improve Machine Translation (MT) system-generated output by following the principle of minimal editing~\cite{bojar-etal-2015-findings, chatterjee-etal-2018-findings}. Along with the shift in the field of MT research- from statistical to neural approaches, research within APE has observed a similar trend- towards neural APE systems~\cite{chatterjee-etal-2018-findings, chatterjee-etal-2019-findings, chatterjee-etal-2020-findings}.


The need for large APE datasets for training neural APE models is addressed by generating artificial triplets \cite{junczys-dowmunt-grundkiewicz-2016-log, negri-etal-2018-escape, freitag-etal-2022-high}. However, unlike real (human post-edited) APE triplets, the artificial post-edits do not follow the \textit{minimality principle}, leading to distributional differences \cite{wei-etal-2020-hw}. Despite training on synthetic data and fine-tuning with real data, current APE systems face over-correction issues, primarily due to the size imbalance between synthetic and real data \cite{chatterjee-etal-2020-findings, bhattacharyya-etal-2023-findings}.

While strategies like optimizing data selection, data augmentation, and model architecture have addressed APE over-correction, mitigating it at the decoding stage remains underexplored~\cite{Carmo2020ARO}. Focusing on other stages limits the applicability across different APE systems. \textbf{Motivated} by this, we propose an over-correction mitigation method using an external Quality Estimation (QE) signal during decoding, applicable to any black-box APE system. Our contribution is:
\begin{itemize}
    \item An over-correction mitigation technique that uses fine-grained word-level QE information to perform constrained decoding. The technique shows improvements of 0.65, 1.86, and 1.44 TER points, respectively, over existing En-De, En-Hi, and En-Mr APE systems.
\end{itemize}

\begin{figure*}[t]
\centering
\includegraphics[width=\linewidth]{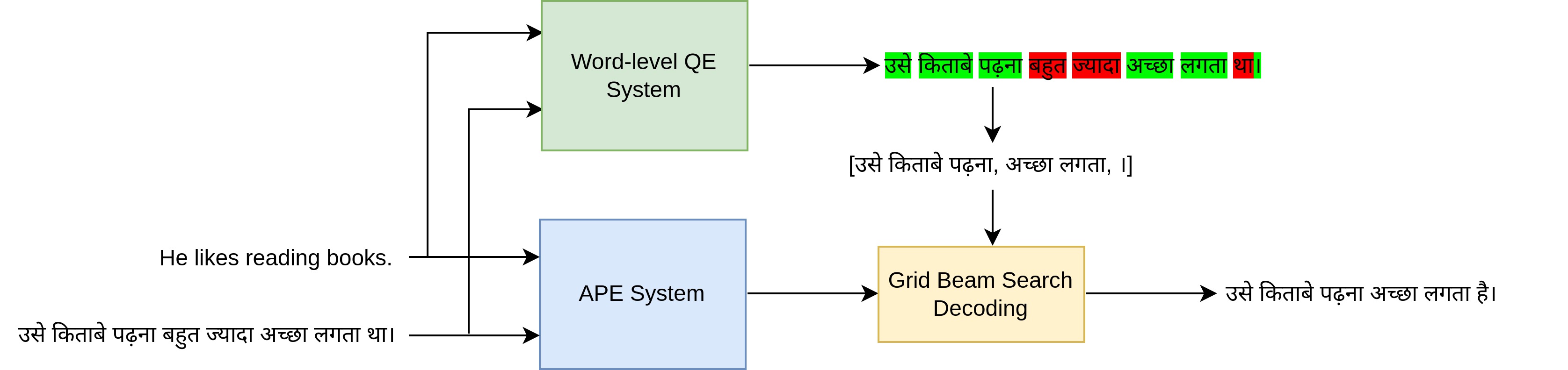}
\caption{An example of the word-level QE-based Grid Beam Search decoding technique used for over English-Hindi APE system. Words marked in green denote word-level QE predicted `OK' tags for them. These correct translation segments (shown in the list) are referred to as \textit{constraints} and are used during the decoding to ensure they appear in the final APE output.}
\label{fig:agbs_diag}
\end{figure*}

\section{Related Work}
There are multiple attempts to curtail the over-correction at different stages of APE development.


\citet{chatterjee-etal-2016-instance, chatterjee-etal-2016-fbk, wang-etal-2021-exploring} focus on data by selecting training samples that may prevent APE from facing the over-correction, augmentation with triplets containing the same translations and post-edits, and weighing training samples with perplexity-based scoring to limit their contribution to learning the APE model.


\citet{junczys-dowmunt-grundkiewicz-2017-exploration} modify their APE architecture using monotonic hard attention to improve translation faithfulness. \citet{chatterjee-etal-2017-multi} use task-specific loss based on attention scores to reward APE hypothesis words present in the original translation. \citet{tebbifakhr-etal-2019-effort} train a classifier to predict post-editing effort and prepend its output to source and translation sequences.


\citet{tan-etal-2017-neural} train separate APE models and use a QE system to rank their outputs. \citet{lee-2020-cross, deoghare-bhattacharyya-2022-iit, yu-etal-2023-hw} mitigate over-correction by reverting to the original translation based on QE speculation. \citet{chatterjee-etal-2018-combining} incorporate word-level QE information into the decoder to guide minimal edits. \citet{deoghare-etal-2023-quality} adopt a multitask approach, jointly training on QE and APE tasks to reduce over-correction.


We find only a few attempts at handling over-correction at the decoding stage. \citet{junczys-dowmunt-grundkiewicz-2016-log} introduce a `Post-Editing Penalty' during decoding to prevent generating tokens not present in the input, applying it in an ensemble framework to one model. \citet{chatterjee-etal-2017-multi} re-rank APE hypotheses based on precision and recall using shallow features like insertions, deletions, and length ratio, rewarding those closer to the original translation. \citet{lopes-etal-2019-unbabels} impose a soft penalty for new tokens not in the inputs. \citet{9721850} experiment with various decoding methods to generate artificial APE triplets.

\section{Methodology}
We use an extension of beam search, called Grid Beam Search \cite{hokamp-liu-2017-lexically}, to perform decoding. While it is originally used for neural interactive-predictive translations and for MT domain adaptation, we adopt the decoding technique for APE. To mitigate the APE over-correction, we explicitly provide information about correct translation segments during the decoding through fine-grained word-level QE signals.

\subsection{Grid Beam Search (GBS)}
Grid Beam Search (GBS) extends the beam search by incorporating lexical constraints into the sequence generation process. Unlike traditional methods that focus purely on maximizing the probability of the output sequence based on the input, GBS allows specific lexical constraints to be mandatorily included in the generated output.

GBS works by structuring the search space into a grid where the rows track the constraints, and the columns represent the progression of timesteps in the sequence. Each cell in this grid holds a set of potential hypotheses, which are candidate output sequences being considered at that point in time. At each timestep, once a new token is generated, it is matched with the start of tokens in the constraint list. If there is a match, the particular constraint is added to the hypothesis. The algorithm evaluates and updates these hypotheses based on whether they comply with the required constraints and how well they fit the model's learned distribution.

The search proceeds by either continuing with a free generation following the standard beam search or by initiating the enforcement of constraints. This balancing act ensures that, by the end of the sequence generation, all specified constraints are included in the translation. Kindly refer to \textbf{Appendix \ref{ap:gbs}} for more details.

\subsection{Word-QE-based Constraints}
A word-level QE system \cite{ranasinghe-etal-2021-exploratory} provides fine-grained information about translation quality by tagging each translation word with an `OK' or `BAD' tag. An `OK' tag indicates the word is a correct translation of some word or phrase in the source sentence. Similarly, a `BAD' tag denotes the word is an incorrect translation and should be deleted or substituted.

We utilize this information to know the correct translation phrases. We first pass the source sentence and its MT-generated translation to the word-level QE system, which provides tags for each token in the translation. We simply consider a set of consecutive tokens with the `OK' tag as a constraint that needs to be present in the APE output. Even though the QE system processes the text at the subword level, we set the `word' to be the smallest unit to be considered as a constraint. Kindly refer to \textbf{Appendix \ref{ap:word-qe}} for details about the word-level QE system.

To summarize, the APE decoding process involves using correct translation segments identified based on the Word-level QE signals and then performing the GBS decoding (Refer Figure \ref{fig:agbs_diag}).

\section{Experimental Setup}

This section details the different experiments undertaken to assess the effectiveness of the proposed decoding technique. We use the same datasets, architecture, data augmentation, and preprocessing and also follow the same training approach as described by \citet{deoghare-etal-2023-quality} for training the APE models to enable direct comparison. \textbf{Appendix \ref{ap:datasets}} details the English-German, English-Hindi, and English-Marathi datasets used for the experiments.

\textbf{Do Nothing} A baseline considering original translations as an APE output.

\textbf{Baseline 1 (Primary Baseline): Standalone-APE + BS}: In this experiment, we train a standalone APE system without any QE data or additionally train the model on QE tasks. The decoding is done using the standard beam search. We consider \textit{Baseline 1} as a \textbf{Primary Baseline}.

\textbf{Baseline 2: QE-APE + BS}: The experiment is an extension of \textit{Baseline 1}. In this experiment, the model is jointly trained on QE and APE tasks as described in \citet{deoghare-etal-2023-quality} by adding QE task-specific heads to the encoders. Similar to \textit{Baseline 1}, this experiment uses the beam search too to perform decoding. This experiment investigates the effectiveness of using word-level QE information during the decoding if the APE model has implicit knowledge of the word-level QE task.

We provide the architecture details and the training approach for both the baselines in \textbf{Appendix \ref{ap:APE Description}} and the hyperparameter information for both APE and QE systems in \textbf{Appendix \ref{ap:hyperparameters}}.

\textbf{Standalone-APE + GBS} In this experiment, we train the APE model as in the \textit{Baseline 1} experiment. However, the decoding is performed using the proposed Word-QE-based GBS decoding technique.

\textbf{QE-APE + GBS} The experiment involves jointly training a model on QE and APE tasks as in the \textit{Baseline 2} experiment. During decoding, instead of standard beam search, the proposed Word-QE-based GBS decoding technique is used.

\section{Results and Discussion}
We perform the experiments on English-German (En-De), English-Hindi (En-Hi), and English-Marathi (En-Mr) pairs, each of which offers a different level of task difficulty due to different linguistic properties, varied amounts of real and synthetic datasets, and `Do nothing' baselines with different complexities. We use TER \cite{snover-etal-2006-study} and BLEU \cite{papineni-etal-2002-bleu} as primary and secondary evaluation metrics, respectively. Kindly refer to Appendix \ref{ap:bleu_scores} for the BLEU scores.

\begin{table}[]
\centering
\resizebox{\columnwidth}{!}{%
\begin{tabular}{lrrr}
\hline
\multicolumn{1}{c}{\textbf{Experiment}} & \multicolumn{1}{c}{\textbf{En-De}} & \multicolumn{1}{c}{\textbf{En-Hi}} & \multicolumn{1}{l}{\textbf{En-Mr}} \\ \hline
\textbf{Do Nothing}                     & 19.06                              & 47.43                              & 22.93                              \\
\textbf{Standalone-APE + BS}                 & 18.91                              & 21.48                              & 19.39                              \\
\textbf{Standalone-APE + GBS (Token)}        & 17.40                              & 19.92                              & 18.48                              \\
\textbf{Standalone-APE + GBS (Word)}         & \textbf{17.74}                     & \textbf{19.43}                     & \textbf{17.31}                     \\ \hline
\end{tabular}%
}
\caption{TER scores on the respective evaluation are set in the Oracle settings when constraint enforcement is done based on initial token or word-based matching.}
\label{tab:tok_or_word}
\end{table}

Table \ref{tab:tok_or_word} compiles the results of experiments geared towards answering whether constraint enforcement should be initiated based on the first token match or the entire word match. These experiments are performed in the oracle setting, meaning ground-truth word-level QE tags are used instead of the word-level QE predicted tags to extract correct translation segments. Better performance in the case of all three pairs when the constraint enforcement is done based on word-based matching indicates the possibility of noise inclusion, as there could be common subword-level prefixes for multiple words that are present across constraints or even non-constraint words. A relatively large difference between \textit{Standalone-APE + GBS (Token)} and \textit{Standalone-APE + GBS (Word)} experiments for En-Hi, En-Mr pairs, and En-De pair hints the noise illusion goes up when target languages are morphologically richer. As we observe consistently better results in the case of \textit{Standalone-APE + GBS (Word)} experiment, further experiments are performed by using word-based matching for enforcing constraints during the GBS decoding.

\begin{table}[]
\centering
\resizebox{\columnwidth}{!}{%
\begin{tabular}{lrrr}
\hline
\multicolumn{1}{c}{\textbf{Experiment}} & \multicolumn{1}{c}{\textbf{En-De}} & \multicolumn{1}{c}{\textbf{En-Hi}} & \multicolumn{1}{l}{\textbf{En-Mr}} \\ \hline
\textbf{Do Nothing}                     & 19.06                              & 47.43                              & 22.93                              \\
\textbf{Standalone-APE + BS}                 & 18.91                              & 21.48                              & 19.39                              \\
\textbf{QE-APE + BS}                    & 18.45                              & 19.75                              & 18.30                              \\
\textbf{Standalone-APE + GBS}                & 18.26                              & 19.62                              & 17.95                              \\
\textbf{QE-APE + GBS}                   & \textbf{18.04}                     & \textbf{19.20}                     & \textbf{17.53}                     \\ \hline
\textbf{Standalone-APE + GBS (Oracle)}       & 17.74                              & 19.43                              & 17.31                              \\
\textbf{QE-APE + GBS (Oracle)}          & 17.50                              & 18.52                              & 16.70                              \\ \hline
\textbf{Greedy}                         & 19.38                              & 20.04                              & 18.73                              \\
\textbf{Sampling}                       & 19.35                              & 19.89                              & 18.46                              \\
\textbf{top-k Sampling}                 & 18.43                              & 19.46                              & 18.18                              \\ \hline
\textbf{\citet{lopes-etal-2019-unbabels}}       & 18.38                              & 19.41                              & 18.16                              \\ \hline
\end{tabular}%
}
\caption{TER scores on the respective evaluation sets in the Oracle and non-oracle settings when different decoding techniques are used.}
\label{tab:main_res}
\end{table}

A comparison between different decoding techniques and the proposed technique is depicted in Table \ref{tab:main_res}. We observe larger improvements with the proposed decoding technique over the standard beam search decoding when the underlying APE system is a standalone system that is not trained for QE tasks. It shows the effectiveness of enforcing the generation of correct translation segments during the decoding.

On the other hand, a smaller difference in improvements between the two techniques when the underlying APE system is jointly trained on QE and APE tasks underlines that the implicit knowledge of the QE tasks helps the model perform APE. Yet, we can conjecture from the better performance with the use of the proposed method over the standard beam search that a loose coupling of QE with APE but with explicit information about the translation segment quality has the potential to improve an APE system developed through the stronger QE and APE coupling.

In both cases, the difference between the proposed technique with oracle and non-oracle word-level QE information underscores the need for better word-level QE systems.

We additionally perform experiments with other popular decoding techniques like greedy, sampling, and top-k sampling for completeness. The \textit{Standalone-APE} model is used in these experiments. The results show that the top-k sampling decoding performs similarly to the beam search decoding. The reported results are with the best \textit{k} values for each pair (En-De: 25, En-Hi: 30, En-Mr: 25) as per empirical observations. We also compare our proposed approach with the work of \citet{lopes-etal-2019-unbabels}, who apply a soft penalty during decoding if APE generates tokens that are not present in either source or translation vocabularies. For this experiment too, we use the standalone APE (Standalone-APE) system. While we observe significant improvements in the case of En-Hi and En-Mr pairs, the technique shows limited gains when compared to the proposed approach, suggesting it is more beneficial to inform APE about what to generate than what not to generate since NMT outputs are usually of high quality and require minimal editing. \textbf{Appendix \ref{ap:deterioration_analysis}} shows how much the decoding technique succeeds in mitigating the over-correction.

The statistical significance test \cite{graham-2015-improving} considering the primary metric (TER) and $p$ being $< 0.05$ shows \textit{Standalone-APE + GBS} experiments show significant gains over their \textit{Standalone-APE + BS} counterparts for all three language pairs. Similarly, improvements through \textit{QE-APE + GBS} over \textit{QE-APE + BS} experiments for all three pairs are significant.

\section{Conclusion and Future Work} 
The proposed decoding technique in this work has demonstrated its effectiveness in enhancing the quality of APE outputs by enforcing the generation of provided correct translation segments during decoding. These segments are extracted with the help of a word-level QE system, which offers fine-grained information about translation quality. Through experiments on three language pairs, En-De, En-Hi, and En-Mr, the technique achieved improvements of 0.87 to 2.28 TER points over baseline APE systems. Notably, the superior performance of standalone APE systems using the proposed decoding method compared to QE-APE systems with traditional beam search decoding underscores the technique's ability to reduce over-correction. This result also suggests that injecting word-level QE information exclusively at the decoding stage is more effective than embedding it implicitly through joint QE and APE training. However, the relatively smaller gains when applying the technique to QE-APE systems imply that incorporating explicit QE information at the decoding stage addresses remaining gaps even after joint training with QE and APE. In the future, we would like to investigate the impact of the quality of a word-level QE system on the proposed decoding technique.

\section{Limitations}
Our technique relies on the availability of a word-level QE system for the language pair of interest. It limits its applicability to a wider set of languages. Furthermore, the results show performance improvements through the proposed technique over the standard beam search are sensitive to the quality of the word-level QE system, which is uncontrolled by nature. The false positives of the word-level QE system will especially lead to the enforcement of the decoding technique to include incorrect translation segments in the output.

\section{Ethics Statement}
Our models for APE and QE are developed using publicly accessible datasets cited in this paper. These datasets have already been gathered and annotated, and this study does not involve any new data collection. Additionally, these datasets serve as standard benchmarks introduced in recent WMT shared tasks. The datasets do not contain any user information, ensuring the privacy and anonymity of individuals. We acknowledge that all datasets carry inherent biases, and as a result, computational models are bound to acquire biased information from them.

\bibliography{anthology,custom}

\begin{thebibliography}{43}
\expandafter\ifx\csname natexlab\endcsname\relax\def\natexlab#1{#1}\fi

\bibitem[{Akhbardeh et~al.(2021)Akhbardeh, Arkhangorodsky, Biesialska, Bojar, Chatterjee, Chaudhary, Costa-jussa, Espa{\~n}a-Bonet, Fan, Federmann, Freitag, Graham, Grundkiewicz, Haddow, Harter, Heafield, Homan, Huck, Amponsah-Kaakyire, Kasai, Khashabi, Knight, Kocmi, Koehn, Lourie, Monz, Morishita, Nagata, Nagesh, Nakazawa, Negri, Pal, Tapo, Turchi, Vydrin, and Zampieri}]{akhbardeh-etal-2021-findings}
Farhad Akhbardeh, Arkady Arkhangorodsky, Magdalena Biesialska, Ond{\v{r}}ej Bojar, Rajen Chatterjee, Vishrav Chaudhary, Marta~R. Costa-jussa, Cristina Espa{\~n}a-Bonet, Angela Fan, Christian Federmann, Markus Freitag, Yvette Graham, Roman Grundkiewicz, Barry Haddow, Leonie Harter, Kenneth Heafield, Christopher Homan, Matthias Huck, Kwabena Amponsah-Kaakyire, Jungo Kasai, Daniel Khashabi, Kevin Knight, Tom Kocmi, Philipp Koehn, Nicholas Lourie, Christof Monz, Makoto Morishita, Masaaki Nagata, Ajay Nagesh, Toshiaki Nakazawa, Matteo Negri, Santanu Pal, Allahsera~Auguste Tapo, Marco Turchi, Valentin Vydrin, and Marcos Zampieri. 2021.
\newblock \href {https://aclanthology.org/2021.wmt-1.1} {Findings of the 2021 conference on machine translation ({WMT}21)}.
\newblock In \emph{Proceedings of the Sixth Conference on Machine Translation}, pages 1--88, Online. Association for Computational Linguistics.

\bibitem[{Bansal et~al.(2013)Bansal, Banerjee, and Jha}]{bansal2013corpora}
Akanksha Bansal, Esha Banerjee, and Girish~Nath Jha. 2013.
\newblock Corpora creation for indian language technologies--the ilci project.
\newblock In \emph{the sixth Proceedings of Language Technology Conference (LTC ‘13)}.

\bibitem[{Bhattacharyya et~al.(2022)Bhattacharyya, Chatterjee, Freitag, Kanojia, Negri, and Turchi}]{bhattacharyya-etal-2022-findings}
Pushpak Bhattacharyya, Rajen Chatterjee, Markus Freitag, Diptesh Kanojia, Matteo Negri, and Marco Turchi. 2022.
\newblock \href {https://aclanthology.org/2022.wmt-1.5} {Findings of the {WMT} 2022 shared task on automatic post-editing}.
\newblock In \emph{Proceedings of the Seventh Conference on Machine Translation (WMT)}, pages 109--117, Abu Dhabi, United Arab Emirates (Hybrid). Association for Computational Linguistics.

\bibitem[{Bhattacharyya et~al.(2023)Bhattacharyya, Chatterjee, Freitag, Kanojia, Negri, and Turchi}]{bhattacharyya-etal-2023-findings}
Pushpak Bhattacharyya, Rajen Chatterjee, Markus Freitag, Diptesh Kanojia, Matteo Negri, and Marco Turchi. 2023.
\newblock \href {https://doi.org/10.18653/v1/2023.wmt-1.55} {Findings of the {WMT} 2023 shared task on automatic post-editing}.
\newblock In \emph{Proceedings of the Eighth Conference on Machine Translation}, pages 672--681, Singapore. Association for Computational Linguistics.

\bibitem[{Bojar et~al.(2015)Bojar, Chatterjee, Federmann, Haddow, Huck, Hokamp, Koehn, Logacheva, Monz, Negri, Post, Scarton, Specia, and Turchi}]{bojar-etal-2015-findings}
Ond{\v{r}}ej Bojar, Rajen Chatterjee, Christian Federmann, Barry Haddow, Matthias Huck, Chris Hokamp, Philipp Koehn, Varvara Logacheva, Christof Monz, Matteo Negri, Matt Post, Carolina Scarton, Lucia Specia, and Marco Turchi. 2015.
\newblock \href {https://doi.org/10.18653/v1/W15-3001} {Findings of the 2015 workshop on statistical machine translation}.
\newblock In \emph{Proceedings of the Tenth Workshop on Statistical Machine Translation}, pages 1--46, Lisbon, Portugal. Association for Computational Linguistics.

\bibitem[{Chatterjee et~al.(2016{\natexlab{a}})Chatterjee, Arcan, Negri, and Turchi}]{chatterjee-etal-2016-instance}
Rajen Chatterjee, Mihael Arcan, Matteo Negri, and Marco Turchi. 2016{\natexlab{a}}.
\newblock \href {https://aclanthology.org/2016.amta-researchers.1} {Instance selection for online automatic post-editing in a multi-domain scenario}.
\newblock In \emph{Conferences of the Association for Machine Translation in the Americas: MT Researchers' Track}, pages 1--15, Austin, TX, USA. The Association for Machine Translation in the Americas.

\bibitem[{Chatterjee et~al.(2016{\natexlab{b}})Chatterjee, C.~de Souza, Negri, and Turchi}]{chatterjee-etal-2016-fbk}
Rajen Chatterjee, Jos{\'e}~G. C.~de Souza, Matteo Negri, and Marco Turchi. 2016{\natexlab{b}}.
\newblock \href {https://doi.org/10.18653/v1/W16-2377} {The {FBK} participation in the {WMT} 2016 automatic post-editing shared task}.
\newblock In \emph{Proceedings of the First Conference on Machine Translation: Volume 2, Shared Task Papers}, pages 745--750, Berlin, Germany. Association for Computational Linguistics.

\bibitem[{Chatterjee et~al.(2017)Chatterjee, Farajian, Negri, Turchi, Srivastava, and Pal}]{chatterjee-etal-2017-multi}
Rajen Chatterjee, M.~Amin Farajian, Matteo Negri, Marco Turchi, Ankit Srivastava, and Santanu Pal. 2017.
\newblock \href {https://doi.org/10.18653/v1/W17-4773} {Multi-source neural automatic post-editing: {FBK}{'}s participation in the {WMT} 2017 {APE} shared task}.
\newblock In \emph{Proceedings of the Second Conference on Machine Translation}, pages 630--638, Copenhagen, Denmark. Association for Computational Linguistics.

\bibitem[{Chatterjee et~al.(2019)Chatterjee, Federmann, Negri, and Turchi}]{chatterjee-etal-2019-findings}
Rajen Chatterjee, Christian Federmann, Matteo Negri, and Marco Turchi. 2019.
\newblock \href {https://doi.org/10.18653/v1/W19-5402} {Findings of the {WMT} 2019 shared task on automatic post-editing}.
\newblock In \emph{Proceedings of the Fourth Conference on Machine Translation (Volume 3: Shared Task Papers, Day 2)}, pages 11--28, Florence, Italy. Association for Computational Linguistics.

\bibitem[{Chatterjee et~al.(2020)Chatterjee, Freitag, Negri, and Turchi}]{chatterjee-etal-2020-findings}
Rajen Chatterjee, Markus Freitag, Matteo Negri, and Marco Turchi. 2020.
\newblock \href {https://aclanthology.org/2020.wmt-1.75} {Findings of the {WMT} 2020 shared task on automatic post-editing}.
\newblock In \emph{Proceedings of the Fifth Conference on Machine Translation}, pages 646--659, Online. Association for Computational Linguistics.

\bibitem[{Chatterjee et~al.(2018{\natexlab{a}})Chatterjee, Negri, Rubino, and Turchi}]{chatterjee-etal-2018-findings}
Rajen Chatterjee, Matteo Negri, Raphael Rubino, and Marco Turchi. 2018{\natexlab{a}}.
\newblock \href {https://doi.org/10.18653/v1/W18-6452} {Findings of the {WMT} 2018 shared task on automatic post-editing}.
\newblock In \emph{Proceedings of the Third Conference on Machine Translation: Shared Task Papers}, pages 710--725, Belgium, Brussels. Association for Computational Linguistics.

\bibitem[{Chatterjee et~al.(2018{\natexlab{b}})Chatterjee, Negri, Turchi, Blain, and Specia}]{chatterjee-etal-2018-combining}
Rajen Chatterjee, Matteo Negri, Marco Turchi, Fr{\'e}d{\'e}ric Blain, and Lucia Specia. 2018{\natexlab{b}}.
\newblock \href {https://aclanthology.org/W18-1804} {Combining quality estimation and automatic post-editing to enhance machine translation output}.
\newblock In \emph{Proceedings of the 13th Conference of the Association for Machine Translation in the {A}mericas (Volume 1: Research Track)}, pages 26--38, Boston, MA. Association for Machine Translation in the Americas.

\bibitem[{Conneau et~al.(2020)Conneau, Khandelwal, Goyal, Chaudhary, Wenzek, Guzm{\'a}n, Grave, Ott, Zettlemoyer, and Stoyanov}]{conneau-etal-2020-unsupervised}
Alexis Conneau, Kartikay Khandelwal, Naman Goyal, Vishrav Chaudhary, Guillaume Wenzek, Francisco Guzm{\'a}n, Edouard Grave, Myle Ott, Luke Zettlemoyer, and Veselin Stoyanov. 2020.
\newblock \href {https://doi.org/10.18653/v1/2020.acl-main.747} {Unsupervised cross-lingual representation learning at scale}.
\newblock In \emph{Proceedings of the 58th Annual Meeting of the Association for Computational Linguistics}, pages 8440--8451, Online. Association for Computational Linguistics.

\bibitem[{Deoghare and Bhattacharyya(2022)}]{deoghare-bhattacharyya-2022-iit}
Sourabh Deoghare and Pushpak Bhattacharyya. 2022.
\newblock \href {https://aclanthology.org/2022.wmt-1.67} {{IIT} {B}ombay{'}s {WMT}22 automatic post-editing shared task submission}.
\newblock In \emph{Proceedings of the Seventh Conference on Machine Translation (WMT)}, pages 682--688, Abu Dhabi, United Arab Emirates (Hybrid). Association for Computational Linguistics.

\bibitem[{Deoghare et~al.(2023{\natexlab{a}})Deoghare, Choudhary, Kanojia, Ranasinghe, Bhattacharyya, and Or{\u{a}}san}]{deoghare-etal-2023-multi}
Sourabh Deoghare, Paramveer Choudhary, Diptesh Kanojia, Tharindu Ranasinghe, Pushpak Bhattacharyya, and Constantin Or{\u{a}}san. 2023{\natexlab{a}}.
\newblock \href {https://doi.org/10.18653/v1/2023.findings-acl.585} {A multi-task learning framework for quality estimation}.
\newblock In \emph{Findings of the Association for Computational Linguistics: ACL 2023}, pages 9191--9205, Toronto, Canada. Association for Computational Linguistics.

\bibitem[{Deoghare et~al.(2023{\natexlab{b}})Deoghare, Kanojia, Blain, Ranasinghe, and Bhattacharyya}]{deoghare-etal-2023-quality}
Sourabh Deoghare, Diptesh Kanojia, Fred Blain, Tharindu Ranasinghe, and Pushpak Bhattacharyya. 2023{\natexlab{b}}.
\newblock \href {https://doi.org/10.18653/v1/2023.findings-emnlp.115} {Quality estimation-assisted automatic post-editing}.
\newblock In \emph{Findings of the Association for Computational Linguistics: EMNLP 2023}, pages 1686--1698, Singapore. Association for Computational Linguistics.

\bibitem[{do~Carmo et~al.(2020)do~Carmo, Shterionov, Moorkens, Wagner, Hossari, Paquin, Schmidtke, Groves, and Way}]{Carmo2020ARO}
F{\'e}lix do~Carmo, D.~Shterionov, Joss Moorkens, Joachim Wagner, Murhaf Hossari, Eric Paquin, Dag Schmidtke, Declan Groves, and Andy Way. 2020.
\newblock \href {https://api.semanticscholar.org/CorpusID:234437292} {A review of the state-of-the-art in automatic post-editing}.
\newblock \emph{Machine Translation}, 35:101 -- 143.

\bibitem[{Feng et~al.(2022)Feng, Yang, Cer, Arivazhagan, and Wang}]{feng-etal-2022-language}
Fangxiaoyu Feng, Yinfei Yang, Daniel Cer, Naveen Arivazhagan, and Wei Wang. 2022.
\newblock \href {https://doi.org/10.18653/v1/2022.acl-long.62} {Language-agnostic {BERT} sentence embedding}.
\newblock In \emph{Proceedings of the 60th Annual Meeting of the Association for Computational Linguistics (Volume 1: Long Papers)}, pages 878--891, Dublin, Ireland. Association for Computational Linguistics.

\bibitem[{Freitag et~al.(2022)Freitag, Grangier, Tan, and Liang}]{freitag-etal-2022-high}
Markus Freitag, David Grangier, Qijun Tan, and Bowen Liang. 2022.
\newblock \href {https://doi.org/10.1162/tacl_a_00491} {High quality rather than high model probability: Minimum {B}ayes risk decoding with neural metrics}.
\newblock \emph{Transactions of the Association for Computational Linguistics}, 10:811--825.

\bibitem[{Graham(2015)}]{graham-2015-improving}
Yvette Graham. 2015.
\newblock \href {https://doi.org/10.3115/v1/P15-1174} {Improving evaluation of machine translation quality estimation}.
\newblock In \emph{Proceedings of the 53rd Annual Meeting of the Association for Computational Linguistics and the 7th International Joint Conference on Natural Language Processing (Volume 1: Long Papers)}, pages 1804--1813, Beijing, China. Association for Computational Linguistics.

\bibitem[{Hokamp and Liu(2017)}]{hokamp-liu-2017-lexically}
Chris Hokamp and Qun Liu. 2017.
\newblock \href {https://doi.org/10.18653/v1/P17-1141} {Lexically constrained decoding for sequence generation using grid beam search}.
\newblock In \emph{Proceedings of the 55th Annual Meeting of the Association for Computational Linguistics (Volume 1: Long Papers)}, pages 1535--1546, Vancouver, Canada. Association for Computational Linguistics.

\bibitem[{Junczys-Dowmunt and Grundkiewicz(2016)}]{junczys-dowmunt-grundkiewicz-2016-log}
Marcin Junczys-Dowmunt and Roman Grundkiewicz. 2016.
\newblock \href {https://doi.org/10.18653/v1/W16-2378} {Log-linear combinations of monolingual and bilingual neural machine translation models for automatic post-editing}.
\newblock In \emph{Proceedings of the First Conference on Machine Translation: Volume 2, Shared Task Papers}, pages 751--758, Berlin, Germany. Association for Computational Linguistics.

\bibitem[{Junczys-Dowmunt and Grundkiewicz(2017)}]{junczys-dowmunt-grundkiewicz-2017-exploration}
Marcin Junczys-Dowmunt and Roman Grundkiewicz. 2017.
\newblock \href {https://aclanthology.org/I17-1013} {An exploration of neural sequence-to-sequence architectures for automatic post-editing}.
\newblock In \emph{Proceedings of the Eighth International Joint Conference on Natural Language Processing (Volume 1: Long Papers)}, pages 120--129, Taipei, Taiwan. Asian Federation of Natural Language Processing.

\bibitem[{Kakwani et~al.(2020)Kakwani, Kunchukuttan, Golla, N.C., Bhattacharyya, Khapra, and Kumar}]{kakwani-etal-2020-indicnlpsuite}
Divyanshu Kakwani, Anoop Kunchukuttan, Satish Golla, Gokul N.C., Avik Bhattacharyya, Mitesh~M. Khapra, and Pratyush Kumar. 2020.
\newblock \href {https://doi.org/10.18653/v1/2020.findings-emnlp.445} {{I}ndic{NLPS}uite: Monolingual corpora, evaluation benchmarks and pre-trained multilingual language models for {I}ndian languages}.
\newblock In \emph{Findings of the Association for Computational Linguistics: EMNLP 2020}, pages 4948--4961, Online. Association for Computational Linguistics.

\bibitem[{Koehn et~al.(2007)Koehn, Hoang, Birch, Callison-Burch, Federico, Bertoldi, Cowan, Shen, Moran, Zens, Dyer, Bojar, Constantin, and Herbst}]{koehn-etal-2007-moses}
Philipp Koehn, Hieu Hoang, Alexandra Birch, Chris Callison-Burch, Marcello Federico, Nicola Bertoldi, Brooke Cowan, Wade Shen, Christine Moran, Richard Zens, Chris Dyer, Ond{\v{r}}ej Bojar, Alexandra Constantin, and Evan Herbst. 2007.
\newblock \href {https://aclanthology.org/P07-2045} {{M}oses: Open source toolkit for statistical machine translation}.
\newblock In \emph{Proceedings of the 45th Annual Meeting of the Association for Computational Linguistics Companion Volume Proceedings of the Demo and Poster Sessions}, pages 177--180, Prague, Czech Republic. Association for Computational Linguistics.

\bibitem[{Lee(2020{\natexlab{a}})}]{lee-2020-cross}
Dongjun Lee. 2020{\natexlab{a}}.
\newblock \href {https://aclanthology.org/2020.wmt-1.81} {Cross-lingual transformers for neural automatic post-editing}.
\newblock In \emph{Proceedings of the Fifth Conference on Machine Translation}, pages 772--776, Online. Association for Computational Linguistics.

\bibitem[{Lee(2020{\natexlab{b}})}]{lee-2020-two}
Dongjun Lee. 2020{\natexlab{b}}.
\newblock \href {https://aclanthology.org/2020.wmt-1.118} {Two-phase cross-lingual language model fine-tuning for machine translation quality estimation}.
\newblock In \emph{Proceedings of the Fifth Conference on Machine Translation}, pages 1024--1028, Online. Association for Computational Linguistics.

\bibitem[{Lee et~al.(2022)Lee, Jung, Shin, and Lee}]{9721850}
Wonkee Lee, Baikjin Jung, Jaehun Shin, and Jong-Hyeok Lee. 2022.
\newblock \href {https://doi.org/10.1109/ACCESS.2022.3154768} {Reshape: Reverse-edited synthetic hypotheses for automatic post-editing}.
\newblock \emph{IEEE Access}, 10:28274--28282.

\bibitem[{Liu(2019)}]{liu2019roberta}
Yinhan Liu. 2019.
\newblock Roberta: A robustly optimized bert pretraining approach.
\newblock \emph{arXiv preprint arXiv:1907.11692}, 364.

\bibitem[{Lopes et~al.(2019)Lopes, Farajian, Correia, Tr{\'e}nous, and Martins}]{lopes-etal-2019-unbabels}
Ant{\'o}nio~V. Lopes, M.~Amin Farajian, Gon{\c{c}}alo~M. Correia, Jonay Tr{\'e}nous, and Andr{\'e} F.~T. Martins. 2019.
\newblock \href {https://doi.org/10.18653/v1/W19-5413} {Unbabel{'}s submission to the {WMT}2019 {APE} shared task: {BERT}-based encoder-decoder for automatic post-editing}.
\newblock In \emph{Proceedings of the Fourth Conference on Machine Translation (Volume 3: Shared Task Papers, Day 2)}, pages 118--123, Florence, Italy. Association for Computational Linguistics.

\bibitem[{Navon et~al.(2022)Navon, Shamsian, Achituve, Maron, Kawaguchi, Chechik, and Fetaya}]{Navon2022MultiTaskLA}
Aviv Navon, Aviv Shamsian, Idan Achituve, Haggai Maron, Kenji Kawaguchi, Gal Chechik, and Ethan Fetaya. 2022.
\newblock Multi-task learning as a bargaining game.
\newblock In \emph{International Conference on Machine Learning}.

\bibitem[{Negri et~al.(2018)Negri, Turchi, Chatterjee, and Bertoldi}]{negri-etal-2018-escape}
Matteo Negri, Marco Turchi, Rajen Chatterjee, and Nicola Bertoldi. 2018.
\newblock \href {https://aclanthology.org/L18-1004} {{ESCAPE}: a large-scale synthetic corpus for automatic post-editing}.
\newblock In \emph{Proceedings of the Eleventh International Conference on Language Resources and Evaluation ({LREC} 2018)}, Miyazaki, Japan. European Language Resources Association (ELRA).

\bibitem[{Oh et~al.(2021)Oh, Jang, Xu, An, and Oh}]{oh-etal-2021-netmarble}
Shinhyeok Oh, Sion Jang, Hu~Xu, Shounan An, and Insoo Oh. 2021.
\newblock \href {https://aclanthology.org/2021.wmt-1.34} {Netmarble {AI} center{'}s {WMT}21 automatic post-editing shared task submission}.
\newblock In \emph{Proceedings of the Sixth Conference on Machine Translation}, pages 307--314, Online. Association for Computational Linguistics.

\bibitem[{Papineni et~al.(2002)Papineni, Roukos, Ward, and Zhu}]{papineni-etal-2002-bleu}
Kishore Papineni, Salim Roukos, Todd Ward, and Wei-Jing Zhu. 2002.
\newblock \href {https://doi.org/10.3115/1073083.1073135} {{B}leu: a method for automatic evaluation of machine translation}.
\newblock In \emph{Proceedings of the 40th Annual Meeting of the Association for Computational Linguistics}, pages 311--318, Philadelphia, Pennsylvania, USA. Association for Computational Linguistics.

\bibitem[{Ramesh et~al.(2022)Ramesh, Doddapaneni, Bheemaraj, Jobanputra, AK, Sharma, Sahoo, Diddee, J, Kakwani, Kumar, Pradeep, Nagaraj, Deepak, Raghavan, Kunchukuttan, Kumar, and Khapra}]{ramesh-etal-2022-samanantar}
Gowtham Ramesh, Sumanth Doddapaneni, Aravinth Bheemaraj, Mayank Jobanputra, Raghavan AK, Ajitesh Sharma, Sujit Sahoo, Harshita Diddee, Mahalakshmi J, Divyanshu Kakwani, Navneet Kumar, Aswin Pradeep, Srihari Nagaraj, Kumar Deepak, Vivek Raghavan, Anoop Kunchukuttan, Pratyush Kumar, and Mitesh~Shantadevi Khapra. 2022.
\newblock \href {https://doi.org/10.1162/tacl_a_00452} {Samanantar: The largest publicly available parallel corpora collection for 11 {I}ndic languages}.
\newblock \emph{Transactions of the Association for Computational Linguistics}, 10:145--162.

\bibitem[{Ranasinghe et~al.(2020)Ranasinghe, Orasan, and Mitkov}]{ranasinghe-etal-2020-transquest-wmt2020}
Tharindu Ranasinghe, Constantin Orasan, and Ruslan Mitkov. 2020.
\newblock \href {https://aclanthology.org/2020.wmt-1.122} {{T}rans{Q}uest at {WMT}2020: Sentence-level direct assessment}.
\newblock In \emph{Proceedings of the Fifth Conference on Machine Translation}, pages 1049--1055, Online. Association for Computational Linguistics.

\bibitem[{Ranasinghe et~al.(2021)Ranasinghe, Orasan, and Mitkov}]{ranasinghe-etal-2021-exploratory}
Tharindu Ranasinghe, Constantin Orasan, and Ruslan Mitkov. 2021.
\newblock \href {https://doi.org/10.18653/v1/2021.acl-short.55} {An exploratory analysis of multilingual word-level quality estimation with cross-lingual transformers}.
\newblock In \emph{Proceedings of the 59th Annual Meeting of the Association for Computational Linguistics and the 11th International Joint Conference on Natural Language Processing (Volume 2: Short Papers)}, pages 434--440, Online. Association for Computational Linguistics.

\bibitem[{Snover et~al.(2006)Snover, Dorr, Schwartz, Micciulla, and Makhoul}]{snover-etal-2006-study}
Matthew Snover, Bonnie Dorr, Rich Schwartz, Linnea Micciulla, and John Makhoul. 2006.
\newblock \href {https://aclanthology.org/2006.amta-papers.25} {A study of translation edit rate with targeted human annotation}.
\newblock In \emph{Proceedings of the 7th Conference of the Association for Machine Translation in the Americas: Technical Papers}, pages 223--231, Cambridge, Massachusetts, USA. Association for Machine Translation in the Americas.

\bibitem[{Tan et~al.(2017)Tan, Chen, Huang, Zhang, Li, and Wang}]{tan-etal-2017-neural}
Yiming Tan, Zhiming Chen, Liu Huang, Lilin Zhang, Maoxi Li, and Mingwen Wang. 2017.
\newblock \href {https://doi.org/10.18653/v1/W17-4776} {Neural post-editing based on quality estimation}.
\newblock In \emph{Proceedings of the Second Conference on Machine Translation}, pages 655--660, Copenhagen, Denmark. Association for Computational Linguistics.

\bibitem[{Tebbifakhr et~al.(2019)Tebbifakhr, Negri, and Turchi}]{tebbifakhr-etal-2019-effort}
Amirhossein Tebbifakhr, Matteo Negri, and Marco Turchi. 2019.
\newblock \href {https://doi.org/10.18653/v1/W19-5416} {Effort-aware neural automatic post-editing}.
\newblock In \emph{Proceedings of the Fourth Conference on Machine Translation (Volume 3: Shared Task Papers, Day 2)}, pages 139--144, Florence, Italy. Association for Computational Linguistics.

\bibitem[{Wang et~al.(2021)Wang, Hardmeier, and Sennrich}]{wang-etal-2021-exploring}
Chaojun Wang, Christian Hardmeier, and Rico Sennrich. 2021.
\newblock \href {https://aclanthology.org/2021.nodalida-main.34} {Exploring the importance of source text in automatic post-editing for context-aware machine translation}.
\newblock In \emph{Proceedings of the 23rd Nordic Conference on Computational Linguistics (NoDaLiDa)}, pages 326--335, Reykjavik, Iceland (Online). Link{\"o}ping University Electronic Press, Sweden.

\bibitem[{Wei et~al.(2020)Wei, Shang, Wu, Yu, Li, Guo, Wang, Yang, Lei, Qin, and Sun}]{wei-etal-2020-hw}
Daimeng Wei, Hengchao Shang, Zhanglin Wu, Zhengzhe Yu, Liangyou Li, Jiaxin Guo, Minghan Wang, Hao Yang, Lizhi Lei, Ying Qin, and Shiliang Sun. 2020.
\newblock \href {https://aclanthology.org/2020.wmt-1.31} {{HW}-{TSC}{'}s participation in the {WMT} 2020 news translation shared task}.
\newblock In \emph{Proceedings of the Fifth Conference on Machine Translation}, pages 293--299, Online. Association for Computational Linguistics.

\bibitem[{Yu et~al.(2023)Yu, Zhang, Yanqing, Zhao, Li, Chang, Li, Miaomiao, Tao, and Yang}]{yu-etal-2023-hw}
Jiawei Yu, Min Zhang, Zhao Yanqing, Xiaofeng Zhao, Yuang Li, Su~Chang, Yinglu Li, Ma~Miaomiao, Shimin Tao, and Hao Yang. 2023.
\newblock \href {https://doi.org/10.18653/v1/2023.wmt-1.85} {{HW}-{TSC}{'}s participation in the {WMT} 2023 automatic post editing shared task}.
\newblock In \emph{Proceedings of the Eighth Conference on Machine Translation}, pages 926--930, Singapore. Association for Computational Linguistics.

\end{thebibliography}

\appendix

\section{Grid Beam Search Decoding \cite{hokamp-liu-2017-lexically}}\label{ap:gbs}

\begin{algorithm}
\caption{Grid Beam Search (GBS)}
\label{alg:gbs}
\begin{algorithmic}[1]
\Procedure{ConstrainedSearch}{model, input, constraints, maxLen, numC, k}
    \State startHyp $\gets$ model.getStartHyp(input, constraints)
    \State Grid $\gets$ initGrid(maxLen, numC, k) \Comment{Initialize beams in grid}
    \State Grid[0][0] = startHyp
    \For{t = 1 to maxLen}
        \For{c = max(0, (numC + t) - maxLen) to min(t, numC)}
            \State $n, s, g \gets \emptyset$
            \For{each hyp $\in$ Grid[t-1][c]}
                \If{hyp.isOpen()}
                    \State $g \gets g \cup$ model.generate(hyp, input, constraints) \Comment{Generate new open hypotheses}
                \EndIf
            \EndFor
            \If{c > 0}
                \For{each hyp $\in$ Grid[t-1][c-1]}
                    \If{hyp.isOpen()}
                        \State $n \gets n \cup$ model.start(hyp, input, constraints) \Comment{Start new constrained hypotheses}
                    \Else
                        \State $s \gets s \cup$ model.continue(hyp, input, constraints) \Comment{Continue unfinished hypotheses}
                    \EndIf
                \EndFor
            \EndIf
            \State Grid[t][c] $\gets k$-argmax$_h \in n \cup s \cup g$ model.score(h) \Comment{k-best scoring hypotheses stay on the beam}
        \EndFor
    \EndFor
    \State topLevelHyps $\gets$ Grid[:][numC] \Comment{Get hypotheses in top-level beams}
    \State finishedHyps $\gets$ hasEOS(topLevelHyps) \Comment{Finished hypotheses have generated the EOS token}
    \State bestHyp $\gets$ argmax$_h \in$ finishedHyps model.score(h)
    \State \Return bestHyp
\EndProcedure
\end{algorithmic}
\end{algorithm}

Algorithm \ref{alg:gbs} describes the steps followed to perform the GBS. In the grid, beams are indexed by variables \( t \) and \( c \). The \( t \) variable denotes the timestep of the search, while \( c \) indicates the number of constraint tokens that are included in the hypotheses for the current beam. It's important to note that each increment in \( c \) corresponds to one constraint token. In this context, constraints form an array of sequences, where individual tokens can be referenced as \( \text{constraints}_{ij} \), meaning token \( j \) in constraint \( i \). The parameter \( \text{numC} \) in Algorithm 1 signifies the total count of tokens across all constraints. We can categorize the hypotheses in beams as (i) \textit{Open} hypotheses, which can start a constraint generation or generate new tokens based on the distribution over the vocabulary provided by the model. (ii) \textit{Closed} hypotheses, which can only generate tokens for the current constraint.

t each search step, the candidates in the beam at \( \text{Grid}[t][c] \) can be generated through three distinct methods:
\begin{itemize}
    \item The open hypotheses from the beam to the left \( (\text{Grid}[t - 1][c]) \) can produce continuations based on the model's distribution \( p_\theta(y_i \mid x, \{y_0, \ldots, y_{i-1}\}) \).
    \item The open hypotheses from both the beam to the left and the one below \( (\text{Grid}[t-1][c-1]) \) can initiate new constraints.
    \item The closed hypotheses from the beam to the left and below \( (\text{Grid}[t-1][c-1]) \) can extend existing constraints.
\end{itemize}

The model described in Algorithm \ref{alg:gbs} provides an interface that includes three functions: \texttt{generate}, \texttt{start}, and \texttt{continue}, which create new hypotheses in each of the three specified manners. It is important to note that the scoring function does not need to be aware of the constraints' presence, although it can include a feature indicating whether a hypothesis is part of a constraint.

The beams located at the top level of the grid (where \( c = \text{numConstraints} \)) hold hypotheses that encompass all constraints. When a hypothesis at this top level produces the end-of-sequence (EOS) token, it can be included in the collection of completed hypotheses. The hypothesis with the highest score from this set is identified as the optimal sequence that satisfies all constraints.

\section{Word-level QE System Description}\label{ap:word-qe}
We approach the word-level QE task as a classification problem at the token level. To predict the word-level labels (OK/BAD), we perform a linear transformation followed by a softmax function on each input token derived from the final hidden layer of the XLM-R model.

\begin{equation}
    \hat{y}_{word} = \sigma(W_{word}^{T} \cdot h_t + b_{word})
\end{equation}

where $t$ indicates the specific token that the model is tasked with labeling within a sequence of length $T$, $W_{word} \in \mathcal{R}^{D \times 2}$ represents the weight matrix, and $b_{word} \in \mathcal{R}^{1 \times 2}$ denotes the bias. The cross-entropy loss function used for training the model is illustrated in Equation~\ref{eq:word_loss}, which resembles the architecture of MicroTransQuest as detailed by \citet{ranasinghe-etal-2021-exploratory}.

 \begin{equation}
\label{eq:word_loss}
     \mathcal{L}_{word} = -\sum^2_{i=1} \Big(y_{word} \odot \log( \hat{y}_{word} ) \Big)[i]
 \end{equation}

\paragraph{Architecture and Training Approach:}
We utilize a transformer encoder to construct the QE models. For generating representations of the input, which consists of the concatenated source sentence and its translation, we use XLM-R~\cite{conneau-etal-2020-unsupervised}. This model has been trained on an extensive multilingual dataset totaling 2.5TB, encompassing 104 different languages, and employs the masked language modeling (MLM) objective, akin to RoBERTa~\cite{liu2019roberta}. Notably, the systems that won the WMT20 shared task for sentence- and word-level QE incorporated XLM-R-based models~\cite{ranasinghe-etal-2020-transquest-wmt2020, lee-2020-two}. Consequently, we implement a similar approach for our word-level QE tasks. To enable token-level classification for word-level QE, we add a feedforward layer atop XLM-R. We train these models based on XLM-R for each language pair using their corresponding word-level QE task datasets. Throughout the training process, the weights of all layers in the model are adjusted.

\section{Datasets}\label{ap:datasets}
For our experiments, we utilize datasets from the WMT21~\cite{akhbardeh-etal-2021-findings}, WMT24\footnote{\href{https://www2.statmt.org/wmt24/qe-subtask3.html}{WMT24 QEAPE Shared Subtask}}, and WMT22~\cite{bhattacharyya-etal-2022-findings} APE shared tasks for English-German, English-Hindi, and English-Marathi, respectively. The datasets for these language pairs comprise 7K, 18K, and 7K real APE triplets, along with 7M, 2.5M, and 2.5M synthetic APE triplets. However, to facilitate a direct comparison with previous studies \cite{deoghare-etal-2023-multi}, we limit the English-German pair to 4M synthetic triplets. Each pair also has a corresponding development set containing 1K triplets for evaluation purposes.

In addition, we incorporate parallel corpora during the APE training process. For the English-Hindi and English-Marathi pairs, we draw upon the Anuvaad\footnote{\href{https://github.com/project-anuvaad/anuvaad-parallel-corpus}{Anuvaad Parallel Corpus}}, Samanantar~\cite{ramesh-etal-2022-samanantar}, and ILCI~\cite{bansal2013corpora} datasets, which each contain approximately 6M sentence pairs. For the English-German pair, we utilize the News-Commentary-v16 dataset from the WMT22 MT task, which consists of around 10M sentence pairs.

For the QE tasks, we also leverage datasets from the WMT21, WMT22, and WMT24 Sentence-level and Word-level QE shared tasks. The English-German QE dataset includes 7K instances for training and 1K for development. The English-Marathi dataset consists of 26K training instances and 1K for development. For English-Hindi, we used the QE-corpus-builder\footnote{\url{https://github.com/deep-spin/qe-corpus-builder}} to gather annotations for translations based on their post-edits.

\section{APE System Description}\label{ap:APE Description}

\paragraph{Architecture:} We design the \textit{Standalone-APE} system using a transformer-based encoder-decoder model. For English-Hindi and English-Marathi, two separate encoders are employed to process the source sentence and its translation, as these languages have different scripts and vocabularies. The outputs from both encoders are fed into two sequential cross-attention layers in the decoder. In contrast, the English-German APE system utilizes a single-encoder, single-decoder architecture due to the shared script and vocabulary between these languages. Here, the source and translation are concatenated with a `<SEP>' tag, and this is encoded by a single encoder, which is passed to a cross-attention layer in the decoder. For both language pairs, the encoders are initialized with IndicBERT~\cite{kakwani-etal-2020-indicnlpsuite} weights.

The only change in terms of the architecture for \textit{QE-APE} is the addition of task-specific (Sentence-level QE and Word-level QE) heads on top of a shared representation layer that takes inputs from the last encoder layers. The representation layer has twice as many neurons for the English-Hindi and English-Marathi pairs compared to the English-German pair, whose size matches that of the final encoder layer. While the \textit{Standalone-APE} is trained only for the APE task with cross-entropy loss, the \textit{QE-APE} is trained jointly for sentence-level sentence-level QE (regression), Word-level QE (token-level classification) and APE tasks, with the Nash-MTL \cite{Navon2022MultiTaskLA} algorithm used for the optimization.

\paragraph{Data Augmentation and Preprocessing}
We enhance the synthetic APE data by incorporating automatically generated phrase-level APE triplets. Initially, we train phrase-based statistical machine translation (MT) systems for both source-to-translation and source-to-post-edit tasks using Moses~\cite{koehn-etal-2007-moses}. In the subsequent step, we extract phrase pairs from both MT systems. APE triplets are then created by aligning the source sides of the extracted phrase pairs. To ensure the quality of the synthetic APE triplets, including the phrase-level ones, we apply LaBSE-based filtering~\cite{feng-etal-2022-language} to eliminate low-quality entries from the synthetic APE dataset. This filtering process involves calculating the cosine similarity between the normalized embeddings of a source sentence and its corresponding post-edited translation, retaining only those triplets with a cosine similarity exceeding 0.91. We obtain approximately 45K phrase-level triplets for the English-Hindi pair, around 50K for English-Marathi, and about 60K for the English-German pair.

\paragraph{Training Approach}
We employ a Curriculum Training Strategy (CTS) for training our APE systems, similar to the approach described by ~\citet{oh-etal-2021-netmarble}. This strategy involves progressively adapting the model to increasingly complex tasks. The steps of the CTS are outlined as follows.

Initially, we train a single-encoder single-decoder model for translating between the source and target languages using the parallel corpus. Next, we enhance the encoder-decoder model for the English-Hindi and English-Marathi APE systems by adding an additional encoder while maintaining the same architecture for the English-German APE. We train the resulting model for the APE task using synthetic APE data in two phases for English-Hindi and English-Marathi and one phase for English-German. In the first phase, the model is trained using out-of-domain APE triplets. The second phase involves training with in-domain synthetic APE triplets. Finally, we fine-tune the APE model with in-domain real APE data.

\begin{table}[]
\centering
\resizebox{\columnwidth}{!}{%
\begin{tabular}{lrrr}
\hline
\multicolumn{1}{c}{\textbf{Experiment}} & \multicolumn{1}{c}{\textbf{En-De}} & \multicolumn{1}{c}{\textbf{En-Hi}} & \multicolumn{1}{l}{\textbf{En-Mr}} \\ \hline
\textbf{Do Nothing}                     & 68.79                              & 38.08                              & 64.51                              \\
\textbf{Standalone-APE + BS}            & 68.91                              & 64.79                              & 68.35                              \\
\textbf{QE-APE + BS}                    & 69.53                              & 66.56                              & 69.72                              \\
\textbf{Standalone-APE + GBS}           & 69.78                              & 66.52                              & 69.99                              \\
\textbf{QE-APE + GBS}                   & \textbf{70.04}                     & \textbf{66.91}                     & \textbf{70.47}                     \\ \hline
\textbf{Standalone-APE + GBS (Oracle)}  & 70.37                              & 66.62                              & 70.68                              \\
\textbf{QE-APE + GBS (Oracle)}          & 70.66                              & 67.72                              & 71.31                              \\ \hline
\textbf{Greedy}                         & 68.42                              & 66.25                              & 69.29                              \\
\textbf{Sampling}                       & 68.43                              & 66.43                              & 69.56                              \\
\textbf{top-k Sampling}                 & 68.35                              & 66.60                              & 69.84                              \\ \hline
\textbf{lopes-etal-2019-unbabels}       & 69.52                              & 66.66                              & 69.89                              \\ \hline
\end{tabular}%
}
\caption{BLEU scores on the respective evaluation sets in the Oracle and non-oracle settings when different decoding techniques are used.}
\label{tab:main_res_bleu}
\end{table}

\section{Training Details}\label{ap:hyperparameters}
Our APE models were trained with a batch size of 32 and allowed a maximum of 1000 epochs, incorporating early stopping with a patience of 5. We utilized the Adam optimizer with a learning rate of 5 x \(10^{-5}\), where \(\beta_1\) is set to 0.9, and \(\beta_2\) is set to 0.997. Additionally, we implemented 25,000 warmup steps. For decoding, we used beam search with a beam size of 5. In the QE experiments, a batch size of 16 was employed, starting with a learning rate of $2e-5$ with using $5\%$ of the training data for warm-up. We also applied early stopping with a patience of 20 steps in the QE and all MTL-based experiments, using WandB for hyperparameter searches. All experiments were conducted on NVIDIA A100 GPUs. The APE model comprises approximately 40 million parameters, with training using the CTS taking around 48 hours, while the QE model contains about 125 million parameters and requires roughly 2.25 hours for training. For preprocessing the English and German datasets, we used the NLTK library\footnote{\url{https://www.nltk.org/}}, and the IndicNLP library\footnote{\url{https://github.com/anoopkunchukuttan/indic_nlp_library}} was used for processing Marathi text. Model training and inference were carried out using Pytorch\footnote{\url{https://pytorch.org/}}. To compute the TER scores, we utilized the official WMT APE and QE evaluation script\footnote{\url{https://github.com/sheffieldnlp/qe-eval-scripts}}, and for BLEU scores, we employed the SacreBLEU\footnote{\url{https://github.com/mjpost/sacrebleu}} library.

\section{BLEU Scores}\label{ap:bleu_scores}

Table \ref{tab:main_res_bleu} reports BLEU scores for the experiments presented in Table \ref{tab:main_res}.

\section{Deterioration Analysis}\label{ap:deterioration_analysis}
\begin{figure}[]
\centering
\includegraphics[width=\linewidth]{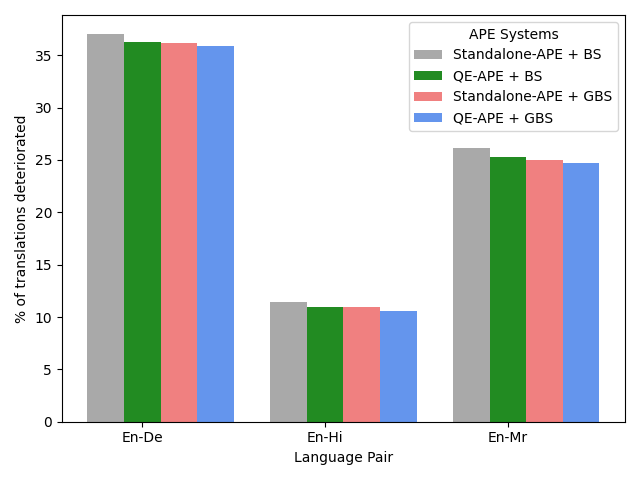}
\caption{Distribution of percentage of different decoding-based APE model outputs with poorer quality than the original translation. \textbf{SA-BS}}
\label{fig:det_plot}
\end{figure}

Figure \ref{fig:det_plot} depicts improvements in the performance through the use of our proposed decoding technique, which points to a reduction in over-correction as the number of APE outputs with poorer quality than the original translation reduces.

\end{document}